# Automated Question Generation on Tabular Data for Conversational Data Exploration

Ritwik Chaudhuri
IBM Research, India

Rajmohan C
IBM Research, India

Kirushikesh DB
IBM Research, India

Arvind Agarwal
IBM Research, India

## ABSTRACT
Exploratory data analysis (EDA) is an essential step for analyzing a dataset to derive insights. Several EDA techniques have been explored in the literature. Many of them leverage visualizations through various plots. But it is not easy to interpret them for a non-technical user, and producing appropriate visualizations is also tough when there are a large number of columns. Few other works provide a view of some interesting slices of data but it is still difficult for the user to draw relevant insights from them. Of late, conversational data exploration is gaining a lot of traction among non-technical users. It helps the user to explore the dataset without having deep technical knowledge about the data. Towards this, we propose a system that recommends interesting questions in natural language based on relevant slices of a dataset in a conversational setting. Specifically, given a dataset, we pick a select set of interesting columns and identify interesting slices of such columns and column combinations based on few interestingness measures. We use our own fine-tuned variation of a pre-trained language model(T5) to generate natural language questions in a specific manner. We then slot-fill values in the generated questions and rank them for recommendations. We show the utility of our proposed system in a coversational setting with a collection of real datasets.

## 1 INTRODUCTION
Exploratory data analysis (EDA) is a critical step for understanding a structured dataset. EDA usually comprises of a set of visualizations, providing a compact view of a subset of data and measures like central tendencies (mean, median, standard deviation etc.) of various columns or a combination of columns in the dataset. Non-technical users may find it challenging to grasp insights from visualizations like histograms, pie charts etc. as these plots might not immediately highlight the key insights. Also, visualizations on more than three columns/attributes are difficult to represent using plots and it is even more difficult to consume it. Hence, visualizations are not enough to summarize the insights from a dataset thoroughly often. Even for technical users, understanding which columns to select next for the EDA process can be difficult, leading to time-consuming analysis due to the vast number of possibilities in the dataset. There are few tools for EDA in some automated fashion like Microsoft Power BI[14], Pandas Profiling[24] or Tableau[21]. They provide an overview of data through plots but have limitations as they don't perform analysis on different slices of the data where relevant insights might reside for a user. Also, user intent is hardly considered in the automated analysis of the data. For example, consider the dataset in Table 1, such tools may provide a histogram of the salary but that will not be sufficient to capture the insight: *"The average salary of employees residing in New York and older than 35 years is $235000"*. This insight is significant because the average salary of employees more than the age of 35 is significantly more than that of the employees who are younger than 35 years. Consider another sample question - *what is the average salary of employees in New York ?* The rationale behind considering such a question is *average salary in New York is significantly more than that of Columbus.* Such insights are possible when analysis is done on relevant and interesting slices of data.

| Employee ID | City | Age | Salary |
|---|---|---|---|
| E01 | New York | 26 | $100000 |
| E02 | New York | 29 | $150000 |
| E03 | Columbus | 29 | $110000 |
| E04 | Columbus | 35 | $210000 |
| E05 | New York | 38 | $250000 |

**Table 1: Salaries of employees in New York & Columbus**

Recently, data exploration via a conversational interface is getting some traction. But there is limited work on automating the data exploration tasks in a conversational setting. In this paper, we propose a system and related methods for data exploration in a conversational setup where a user may upload a dataset for analysis and the system generates interesting questions in natural language based on relevant slices of the data. The system generates unique and interesting questions in natural language based on the dataset, helping users focus on relevant sections of the dataset for deeper analysis, and these questions become more relevant and personalized with each iterative selection in the conversational setup.

The proposed system is aimed for conversational setup such as chatbots where a user uploads a dataset and the system generates questions on relevant slices of data. One advantage of generating these questions in a natural language is that a question may be formed using a large number of columns and yet the question is easily understandable. Another advantage is that the user may not need to have any prior knowledge about the data before analyzing it. Moreover, based on feedback obtained from the users the system can incrementally improve its recommended questions. We propose a system that uses a set of standard interestingness measures and slot filling techniques along with a fine-tuned language model(pre-trained) for achieving this. As per our limited knowledge, there is no existing system that is specifically proposed for generating interesting natural language questions, specifically aggregate questions for structured data exploration in a conversational setting.

## 2 LITERATURE REVIEW

Data exploration and analysis is a challenging and time-consuming task hence there have been a lot of work in automating and improving the processes. There are many tools like Tableau[21] and Microsoft Power BI[14] that aid in exploratory data analysis by producing various visualization plots of the data. There have been various recommender systems proposed for EDA that are aimed at assisting users in choosing the next best exploratory step to perform or suggesting dataset slices that are likely to be of interest. To that end, various interestingness measures have been proposed in the literature.

Data-driven EDA systems define a notion of interestingness and use it to rank EDA operations and the utility of their results. Some early works like [10, 22] have looked at data cubes, roll-ups, drill-downs etc. in traditional databases and OLAP settings. Some works[26, 28] have looked at it from a visualization perspective. Specifically, [26] talks about visualizations that demonstrate a larger deviation from the original dataset as interesting ones. Some systems[1, 6, 25] utilize logs from past exploration sessions or of the current session itself to generate recommendations for exploratory operations to be performed next. [16] proposed a hybrid approach combining data-driven and log-based approaches.

There have also been some machine learning-based approaches for modelling user interests in order to recommend EDA operations. [16] considers a set of interestingness measures, and formulates a multi-class classification problem for selecting interestingness measure dynamically that best captures user interests in every step of an ongoing EDA session. [9] proposes an active learning approach to get feedback on whether the presented tuples are interesting and use it to model user's interest incrementally. [13] utilizes user-annotated data visualizations for training and builds a ranking model that is able to assess the quality of data visualizations and decide, given two data visualizations, which one is more interesting. [2, 15] employs deep reinforcement learning techniques to auto-generate exploratory sessions given a dataset as input which can then be presented to users in some form like a python notebook.

Conversation data exploration over datasets is a fairly recent phenomenon. There have been some early works from the research community in the related area of natural language interfaces to interact with databases. For example, [12] takes natural language sentences from the user and generates a query in a technical language like SQL, presenting its usability and limitations. More recently, [4] presents a no-code platform for binding conversational flows to relational data sources visually. [20] presents an interactive paradigm for the rapid prototyping of chatbots for data exploration. [5] presents an approach for generating chatbots to query Open Data sources published as web APIs.

[3] presents a data-driven design paradigm for building conversational interfaces for data exploration. It exploits the properties of data models and proposes schema annotations to enable the generation of conversation paths for the exploration of database tables. [7] presents a conversational approach for querying multidimensional data, that captures users' intentions and links them to the database schema metadata using a bot development framework. It relies on database schema vocabulary for detecting user intentions and establishing parameters for query execution.

Question generation on tables aims to generate questions from given tabular and associated textual data. Initial approaches have been largely based on syntax rules or templates. Later, supervised neural models have been tried out for the same. These supervised methods generally require large amounts of human-written questions for training. While many works focused on fact-based single-hop questions[27], there have also been some works which have looked at more complex multi-hop questions[18] as well. After pre-trained language models became more popular, they have been used widely for question generation tasks[23]. [19] presents an answer-aware question generation from tables along with text using a transformer-based model. But it is based on the assumption that the answer is known and then fact-based questions are generated.

## 3 INTERESTING NL QUESTION GENERATION FOR EDA

### 3.1 Problem Statement

Given a dataset $D$ with $m$ column headers denoted by $C_1, C_2, \ldots, C_m$, the goal is to generate relevant questions spanning over a subset of columns $\{C_{j_1}, C_{j_2}, \ldots, C_{j_r}\}, 1 \leq r \leq m$ where without loss of generality we assume $j_1 < j_2 < \ldots < j_r$ and $j_i \in \{C_1, C_2, \ldots, C_m\}$. The questions are generated based on relevant slices from the sub-dataset formed by $[C_{j_1} : C_{j_2} : \ldots : C_{j_r}]$. After generation and ranking of questions, the system improves relevance of the questions in subsequent iterations based on the user's past interactions.

### 3.2 The Pipeline Architecture

Our proposed system comprises of mainly three steps: (1) Performing EDA on user uploaded dataset (2) Generating questions based on the significant results obtained from EDA (3) Feedback provided by the user to the system to fine-tune the questions in the next iterations. This entire pipeline is shown in Figure 1. A user uploads a dataset or searches for a dataset within an existing catalog. Once the dataset is uploaded or discovered by the user, using all possible subsets of the columns, various slices are created on the dataset and for each slice, an interestingness score is measured using the metrics provided in [17] along with coefficient of variation, the standard deviation for the numerical columns, correlation among pairs of numerical columns etc. For each subset of columns, the slice with the maximum interestingness score is selected. Using the slice and the columns a natural language question is formed by the system. The entire process is repeated until all the subsets of columns have been covered and then all the questions are ranked based on the interestingness score. The top K questions are shown to the user. This begins the interactive session between the user and the system. The user may provide a column header of interest or select some relevant questions. Then the system weighs the columns that appeared in the user-selected questions and the column headers provided by the user higher and generates questions based on those combinations of column headers. It again ranks the questions before showing the top K questions to the user in the next iteration. This interactive session continues until the user halts it.

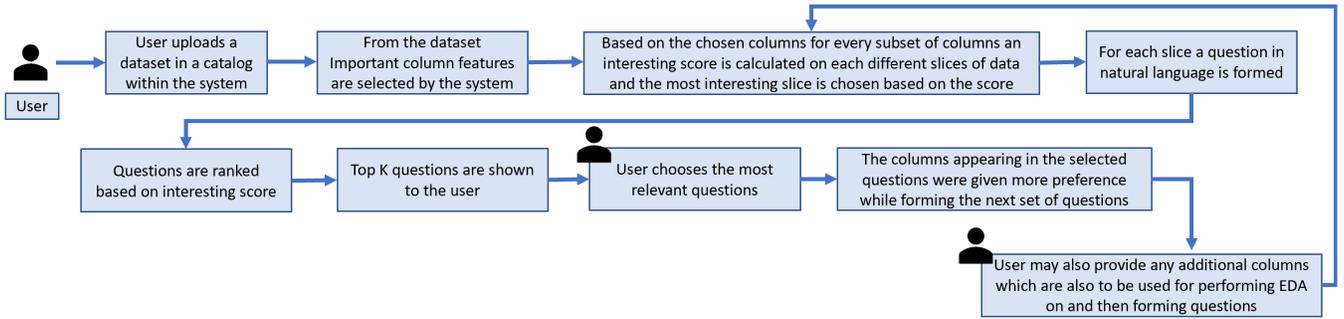

Figure 1: Pipeline for question generation through EDA along with feedback

## 4 DETAILED METHODOLOGY

The entire methodology for generating and ranking of questions given a dataset can be divided into four broad steps (1) Determining interesting columns (2) Determining relevant slices of sub-dataset (3) Question formation and slot-filling (4) Ranking of generated questions. There are certain set of *operators* that are considered for columns of the dataset. For example, consider the question, *"What fraction of employees has a job title as Software Developer ?"* involving the column *job titles*. In the question, the operator is *fraction* which is essentially applied to the proportion of instances for employees with the job title software developer as compared to all the job titles. In this case, a single operator is applied to a single column. Another example question can be, *"What is the average salary of the employees with a salary above $6000 ?"*. Here essentially two operators *average* and *above* are applied on the numeric column *salary*. Hence two operators are applied to a single column. In another example, *"What is the average salary of employees belonging to top 4 job titles ?"*, involving the columns *salary* and *job titles*, the operator *average* is applied on the numerical column *salary* and the other operator *top K (here K = 4)* is applied on the categorical

column *job titles*. So, in this case, each operator is applied on each column. Consider the question, *What is the average salary above age 45 among females ?*, here the operator *average* is applied on the column *salary* and the operator *above* is applied on the column *age*. In a generic aggregated question there can be multiple operators applied multiple times on several columns. In our work, we have considered the operators *Average, Minimum, Maximum, More than, Less than, Above, Below, Top K percent, fraction, Total, Majority, Minority, Among, Missing, Outlier, After, Before, Within, On* applied across numerical, categorical and date-type columns.

We now describe all the four steps of the methodology in detail in the following four subsections when a user uploads a dataset $D$ or fetches a dataset $D$ from the catalog.

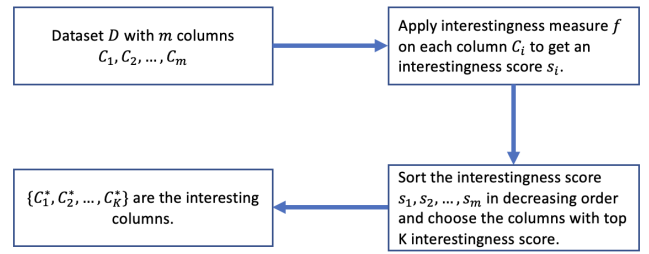

Figure 2: Method to determine interesting columns within the input dataset

### 4.1 Determine interesting columns

Consider the dataset $D$ with $m$ columns namely $C_1, C_2, \ldots, C_m$. On each column $C_i, 1 \leq i \leq m$ an interestingness score function $f : C_i \rightarrow R$ is applied which generates an interestingness score for the column $C_i$. The function $f$ can be any of the popular functions such as Entropy, Unlikeability, Peculiarity, Fisher's score, Chi-squared, and Correlation available in the literature. Then, top $K, 1 \leq K \leq m$ columns, $\{C_1^*, C_2^*, \ldots, C_K^*\}$, are chosen based on the interestingness score. The entire pipeline for determining interesting columns in an input dataset is given in Figure 2. Every subset of the $K$ columns is considered as a sub-dataset and within that sub-dataset the relevant slices are searched for.

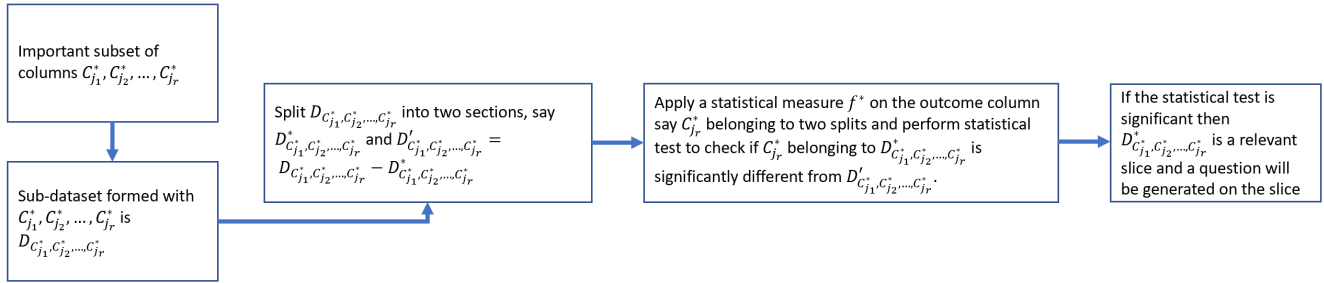

Figure 3: Method to determine relevant slices within the sub-dataset

## 4.2 Determine relevant (important) slices within a sub-dataset

Consider a subset of $r$ columns $\{C^*_{j_1}, C^*_{j_2}, \ldots, C^*_{j_r}\}$, where without loss of generality assume $j_1 < j_2 < \ldots < j_r$ and $j_i \in \{1, 2, \ldots, K\}$, $\forall i, 1 \leq i \leq r$ and the sub-datasets created by the $r$ columns are denoted by $D_{\{C^*_{j_1}, C^*_{j_2}, \ldots, C^*_{j_r}\}} = [C^*_{j_1} : C^*_{j_2} : \ldots : C^*_{j_r}]$. Among $r$ columns, one column is fixed on which a centrality measure, such as average, median, mode, standard deviation etc. denoted by $f_*$ is applied. For example, in the question, *"What is the average salary above age 45 among females ?"*, the fixed column is *salary* on which the measure $f_*$ average is computed. Now, the sub-dataset is sliced based on the values within the remaining $r-1$ columns, specifically for this example columns *age* and *gender*. Hence for each slice the average salary is compared against the salary of the rest of the sub-dataset without the competing slice and checked if it is significantly different from the salary belonging to rest of the sub-dataset. If the difference is significant then the specific slice is taken as a relevant/important slice within the sub-dataset. The entire pipeline for finding the relevant slices of data is given in Figure 3. As shown in Figure 5 the column Salary is the fixed column and the rest of the columns age, gender are the remaining 2 columns. Based on the combination of values within age and gender the sub-dataset is sliced as shown in Figure 5 bordered with a red box. The data bordered with the green box is the remaining sub-dataset. The choice of the measure $f_*$ varies based on the column on which it is applied being a numerical column, categorical column or a date-type column.

If the fixed column is numerical then the function $f_*$ can be any measure of central tendency for which various statistical tests such as two-sample tests can be done to check if the sample of the numerical column within the slice is significantly different from the rest. If the slice is significantly different from the rest, then it is considered to be one of the relevant slices of the data. Based on the structure of the slice, on $r-1$ columns the operators $\{o_1, o_2, \ldots, o_{r-1}\}$ where $o_i \in \{$before, after, in, more than, less than, among, within$\}$ operators are applied on each of the $r-1$ columns depending on whether the column is numerical or categorical or a date-type. For the fixed column based on the function $f_*$ the operator $o_r$ can be *average, median, quartiles, quantiles, standard deviation, trend, coefficient of variation etc.*.

If the fixed column is categorical then the function $f_*$ can be a majority or minority which essentially determines the class of the fixed column within a slice. The slice is created based on the combination of values in the rest of the $r-1$ columns. The function $f_*$ can also be *fraction of elements within a class*. Based on the slice, as stated earlier the operators $\{o_1, o_2, \ldots, o_{r-1}\}$ applied on each of the $r-1$ columns can be *before, after, within, more than, less than, among*. The operator $o_r$ on the fixed column can be a majority, minority, fraction etc. Once the relevant slice $D^*_{\{C^*_{j_1}, C^*_{j_2}, \ldots, C^*_{j_r}\}} \subseteq D_{\{C^*_{j_1}, C^*_{j_2}, \ldots, C^*_{j_r}\}}$ is determined on the sub-dataset formed by columns $\{C^*_{j_1}, C^*_{j_2}, \ldots, C^*_{j_r}\}$ and operators $\{o_1, o_2, \ldots, o_r\}$, Step 4.3 is executed next.

## 4.3 Question formation and slot filling

In this step, we generate a question based on the slice of the data $D^*_{\{C^*_{j_1}, C^*_{j_2}, \ldots, C^*_{j_r}\}}$, columns $\{C^*_{j_1}, C^*_{j_2}, \ldots, C^*_{j_r}\}$ and operators $\{o_1, o_2, \ldots, o_r\}$. This step however is divided into two stages (a) Question formation based on the column headers within the sub-dataset. (b) Slot-filling of generated questions with numbers.

*4.3.1 Question formation based on the column headers within the sub-dataset.* We use pre-trained *text-to-text* models such as *T5-base* for the questions generation. $T5 - base$ is a popular pretrained language model. It can be fine-tuned for a specific task. In fact, for aggregated question generation since there is no training dataset available we hand-curated a training corpus comprised of multiple datasets with (i) a title of the dataset if there is any (ii) a short sentence mentioning the names of the columns contained within the dataset (iii) a dictionary which maps the name of the column header to operator applied on it. The corresponding output was a question generated based on the names of column headers within the sub-dataset. The I/O structure is shown in Figure 4. The system learns how to use numerical operators such as *average, median, mode, standard deviation, maximum, minimum etc.* on the numerical columns and categorical operators such as *most, least etc.* on categorical columns. Also, operators such as *within, from-to etc.* are applied on date-type columns. It is not mandatory to use $T5$ as the only model since our methodology can easily be extended to other pre-trained text-to-text models. Let the tuned text-to-text model through few-shot learning be denoted by $M$. The input parameters for the model $M$ are as shown in Figure 4. For example, if the following inputs *(i) Title: A dataset with age, gender, location and salary of employees (ii) Description: The dataset contains the age of employees, gender of employees, location of employees, the salary of employees (iii) Dictionary: age: above, female: among, location:*

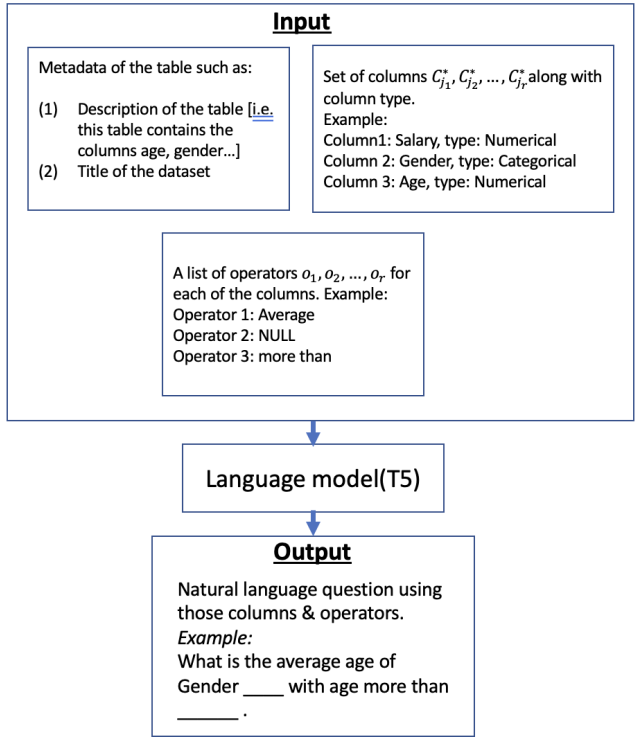

Figure 4: Input and Output for fine-tuning the pre-trained T5 model

in, salary: average is passed through the model $M$ during question generation (testing), then the question generated is: What is the average salary above age ______ among females ? The slot-filling is essentially done in Stage 2 given in Subsection 4.3.2.

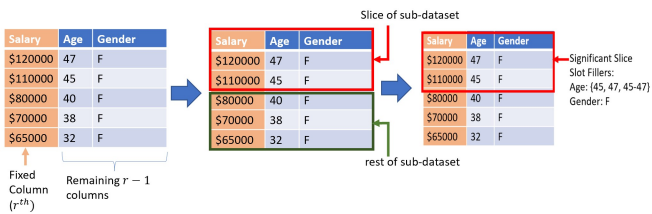

Figure 5: An example dataset with a significant slice

*4.3.2 Slot-filling of generated questions with numbers.* The slot-filling method described here is one of the many techniques that can be used for slot-filling.

From stage 1 described in Subsection 4.3.1, let the generated question be denoted by $q$ with multiple blanks which are to be slot-filled. Based on the relevant slice of data denoted by $D^*_{\{C^*_{j_1}, C^*_{j_2}, ..., C^*_{j_r}\}}$ we obtain range, maximum and minimum as potential slot-filler values of each numerical column. For each categorical column the categories within the slice $D^*_{\{C^*_{j_1}, C^*_{j_2}, ..., C^*_{j_r}\}}$ are chosen as potential slot-fillers. Based on what each of $\{o_1, o_2, \ldots, o_{r-1}\}$ the operators are for each column the appropriate slot-filler value is chosen from the pull of potential slot-filler values. For the example *"What is the average salary above age ______ among females ?"*, the potential slot fillers are shown in the Figure 5. However, since the operator that appears before age is above, the value chosen is 45. Hence, when the question obtained from stage 1 was *"What is the average salary above age ______ among females ?"* when passed through stage 2, the sentence becomes "What is the average salary above age 45 among females ?" and the slot is filled with the number obtained from the sliced sub-dataset.

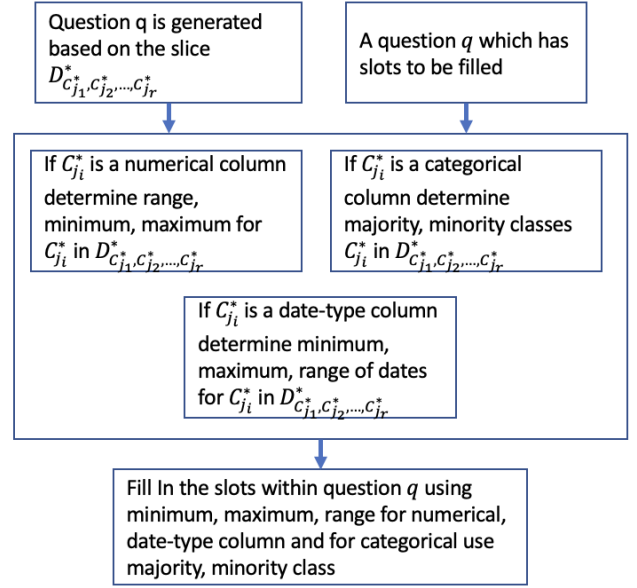

Figure 6: Methodology followed to slot-fill the blanks

Once all the questions are generated based on various slices of several sub-datasets and the slots are filled in, the questions are ranked. For the initial iteration the ranking of the questions are done based on the interestingness scores. However, based on user's feedback the re-ranking of questions are re-done in the subsequent iteration. The methodology for ranking the questions is provided in step 4 described in Subsection 4.4.

### 4.4 Ranking of generated questions

Let us assume from Step 3, given in Subsection 4.3, the set of all questions generated are $Q = \{q_1, q_2, \ldots, q_n\}$. For the question $q_i, 1 \leq i \leq n$ the corresponding interesting score is denoted by $s_i$. In the initial iteration, the questions are ranked in the decreasing order of $s_i$. However, at the beginning of iteration 2 we assume the user is equally likely to choose questions with any of the columns $C^*_i, 1 \leq i \leq K$. Hence we keep a counter denoted by $T_i$ initialized at 1 to denote the number of times a column appeared within the questions selected by the user. Whenever the user chooses a question with the column $C^*_i$ then $T_i$ is increased by 1. The probability for choosing column $C^*_i$ is $p_i = \frac{T_i}{\sum_{j=1}^{K} T_j}$. Starting from Iteration 2, for each question with columns $C^*_{j_1}, C^*_{j_2}, ..., C^*_{j_r}$ the probability of the columns appearing in those questions are computed as $\prod_{l=1}^{r} p_{j_l}$.

The questions are then ranked in decreasing order of probability. Among the questions with the same probability, those are ranked in decreasing order of the interestingness score. Again for further iterations the probabilities of each of the columns change based on the user's selections of questions and hence the ranking of the questions keeps on changing. Step 4 continues recursively based on the user's feedback obtained after each iteration.

## 5 SYSTEM IMPLEMENTATION

### 5.1 Dataset

We picked a set of Kaggle[11] datasets primarily belonging to Finance, Retail and Education domains then created a bunch of interesting aggregate questions for each dataset manually. With that we prepared a new dataset called agg-questions dataset that has the basic metadata of the input dataset columns along with our manually curated aggregate questions.

The metadata considered includes

- table name
- table description
- column name(s)
- column type(s)
- operators

Figure 7 shows operators used in the training data for the three categories of columns that we consider namely numerical, categorical and date columns. In addition, Trend, and Seasonality operators are also considered. Please note that the terms dataset and table are used interchangeably in this section to convey the same thing.

| Numerical | Categorical | Date |
|---|---|---|
| Average, Minimum, More than, Filter, Top K percent, etc. | Fraction, Total, Majority, Missing & Minority | After, Before, On, Missing & Filter |

**Figure 7: Operators used for different column types**

The questions considered can be classified into three categories.

- Single_column-single_operator: Single column of the table along with single operator is used. For example, 'column=age and operator=average' produces 'What is the average age of customers?'
- Single_column-two_operator: Single column of the table along with two operators is used. For example, *column = discount and operator1 = average, operator2 = top* produces 'What is the average discount of top k percent products ordered?'
- Two_column-two_operator: Two columns of the table along with two operators are used. For example, *column=age and operator1 = average, operator2 = filter* produces 'What is the average age of customers with bank balance above X ?'

For single-column questions, 47% of them are on numerical columns and 35% of them are on categorical columns and the remaining are on date columns. Similarly, for two-column questions, the primary column type combinations are Cat-Num(32%), Cat-Cat(19%), Num-Num(18%), Num-Cat(15%) and the remaining are some combination of date columns with categorical/numerical columns. The total number of questions in this manually curated dataset (agg-questions) is around 500.

| Customer ID | Name | Marital Status | Dependent Count |
|---|---|---|---|
| 12112 | ABC | Single | 2 |
| 12113 | XYZ | Married | 4 |
| 12337 | EFG | Divorced | 1 |

- What fraction of customers have marital status as Divorced?
- How many customers have dependent count more than 6?
- What is the maximum dependent count of customers?

| Employee ID | Employee Name | Job Title | Termination Date |
|---|---|---|---|
| 12112 | ABC | Senior Developer | 10/21/2021 |
| 12113 | XYZ | Architect | 11/01/2021 |
| 12337 | EFG | Architect | 13/01/2021 |

- What fraction of employees have job title as Architect and Termination Date between 01/01/2021 and 31/03/2021?
- Which job title type has minimum number of employees leaving after 01/01/2021?

**Figure 8: Questions generated by the system on given tables**

### 5.2 Model

We have used the T5 pre-trained language model(t5-base) from Huggingface model repository and fine-tuned it on this custom dataset(agg-questions) as explained in Section 4. We have experimented with this fine-tuned model as a part of the whole system and the results obtained are discussed in the next section.

### 5.3 Output

Figure 8(top) shows a snapshot of a table called Customer_details that is given as input to the system. The top 3 questions generated by the system are shown just below it. These questions are framed on some important attributes of the table and subsequent slot-filling ensures important slices of the data are captured by these questions.

Figure 8(bottom) shows a snapshot of Employee_offboarding table and at the bottom of the table, some sample questions generated by the system are shown. The column selections and slot-fillings are done based on the methodology explained in section 4.

In Table 2, some of the generated questions by the system on a set of Kaggle datasets from diverse domains are shown. Next we demonstrate a user interaction with the system for exploratory data analysis.

### 5.4 Exploratory analysis using the system

The user interaction with the system begins when the user uploads a dataset for analysis. Figure 9a is the introduction screen of the Question Generation Toolkit which the user finds when the application is launched. The user can see the catalog on the left pane with all the datasets that are within the catalog. As shown in Figure 9a the user may upload a new dataset or the user can continue with one of the previously saved sessions. If the user chooses the option to upload a dataset then as shown in Figure 9b user needs to choose between two more options, (i) upload data from a local machine or (ii) upload data from Catalog. In case the user chooses the option for uploading a dataset from catalog the user may navigate through the datasets provided in the catalog as shown in the left pane. From Figure 9c it can be observed that the user chose the option to upload a dataset from a local machine. Also, in Figure 9c it can be seen that the user uploaded a dataset named *resident_expenditure.xlsx*. Also from Figure 9c it can be observed that through UI the user can provide a limit on the number of questions to be generated. As shown in Figure 9c the user chose a limit of 500 questions to be generated. The system then generates the questions. In Figure 9d it can be seen that the user can choose either of the options of searching questions based on keywords or the user can see all the generated questions. In case the user chooses to see all the questions then the system shows all the questions to the user as depicted in Figure 9d. The user may select any of the questions. As shown in Figure 9d the user ended up choosing the red highlighted question which is based on the city *Chapel Hill, North Carolina*. Hence in the next iteration, we observe from Figure 9e that more questions have been shown to the user based in Chapel Hill, North Carolina. This is because based on the user feedback from the past iterations the system recursively ranks the questions and in the current iteration shows the most relevant question to the user. Also, the entire user session is saved so that the user may again revisit this dataset and can start from wherever the user stopped in the last session. As shown in Figure 9d the user may also choose to search for questions by writing keywords which the system auto-fills with the relevant questions. As can be observed from Figure 9f, within the yellow box the user entered the words such as *In Chapel Hill, What is the average salary* and the system recommended the auto-filled questions given below the yellow box. The user may choose any of the recommended questions (the red highlighted question is selected by the user) and then the system re-ranks the questions for the current iteration based on user-chosen questions in the past iterations. The entire system has been created so that the user can interact with the system through a very easy-to-understand UI and yet have enough freedom to guide the analysis towards a certain direction by providing feedback to the system iteratively.

| Dataset | Generated questions |
| --- | --- |
| auto-insurance-claims-data | What fraction of claims had age more than 60? <br> Which zip had the most number of claims when policy_bind_date is between 01-01-2006 and 31-03-2006? <br> What is the average umbrella_limit when policy_state is OH? <br> What fraction of incident_severity is major_damage when collision_type is rear collision? <br> How many claims are there with auto_year before 2005? |
| bank-marketing-dataset | What is the major education_method when balance greater than 5000? <br> What fraction of people have contact method as unknown? <br> Which job type has minimum defaulters? <br> What is the average balance of the bottom 10% of the people by balance? <br> What is the total number of people with marital as single and housing as yes? |
| top-play-store-games | Which title has the highest total ratings? <br> Which title had maximum growth in the last 30 days? <br> What is the total number of games with install milestones greater than 100 M? <br> Which games have 1-star ratings of more than 100000? <br> Which game category has the maximum paid games? |
| melbourne-housing-market | Which housing type has a minimum average price? <br> What is the major suburb when the date sold is after 01/01/2017? <br> What type of house has the lowest average price in Albion? <br> How many sellers are there in Alphington suburb? <br> What is the average price in Alphington when the number of rooms is more than 5? |
| edx-courses | Which course type has maximum enrolled? <br> What is the average course length of computer science subject? <br> What are the cheapest courses in Data Analysis & Statistics? <br> How many courses are there in Business & Management with Espanol as language? <br> Which intermediate courses have more than 4 instructors? |
| car-sales | Which car manufacturer has maximum average sale price ? <br> Which is the majority vehicle type for car manufacturer Audi? <br> Which model car has the highest average yearly resale value? <br> Who launched the most cars between 01/01/2011 and 31/01/2011? <br> Which intermediate courses have more than 4 instructors? <br> What is the maximum engine size of cars with lengths less than 175? |

Table 2: Kaggle datasets and generated questions on those datasets

## 6 DISCUSSION

Although the system generates meaningful and valid questions in most cases, sometimes the question generated can be invalid. To

| Toolkit for Generating Aggregated Questions Based on a Dataset | | |
|---|---|---|
| Catalogs | Upload Dataset | Continue on Saved Session |
| 📁 Weather data<br>📁 Sports data<br>📁 Retail Data | | |

**(a) The introduction screen of the Toolkit**

| Toolkit for Generating Aggregated Questions Based on a Dataset | | |
|---|---|---|
| Catalogs | Upload Dataset | Continue on Saved Session |
| 📁 Weather data<br>📁 Sports data<br>📁 Retail Data | 📄 Upload data from local machine<br>📁 Upload data from Catalog | |

**(b) Upload Dataset from Local Machine option is Chosen**

| Toolkit for Generating Aggregated Questions Based on a Dataset | | |
|---|---|---|
| Data Uploaded | Search Questions | Show all questions |
| 🗎 resident_expenditure.xlsx | ⧖ `Wait while questions are being generated` | |
| | Number of Questions to be Generated | 500 |

**(c) Putting a limit on the questions generated**

| Toolkit for Generating Aggregated Questions Based on a Dataset | |
|---|---|
| Data Uploaded | Show all Questions |
| 🗎 resident_expenditure.xlsx | 1. What is the average expenditure of residents with more than 3 members in New York ?<br>2. What is the average expenditure of residents with salary more than $120000 ?<br>3. Residents of which state has lowest monthly expenditure ?<br>4. **What is the average salary of residents with one house in Chapel Hill, North Carolina ?** |

**(d) All the questions are shown, red highlighted question is chosen by user**

| Toolkit for Generating Aggregated Questions Based on a Dataset | |
|---|---|
| Data Uploaded | Show all Questions |
| 🗎 resident_expenditure.xlsx | 1. **What is the median salary of residents in Chapel Hill, North Carolina ?**<br>2. What fraction of employees in Chapel Hill, North Carolina has income more than $60000 ?<br>3. What is the average expenditure of residents in Chapel Hill, North Carolina with a family size more than 5 ? |

**(e) Revised questions based on user's feedback**

| Toolkit for Generating Aggregated Questions Based on a Dataset | |
|---|---|
| Data Uploaded | Search Questions |
| 🗎 resident_expenditure.xlsx | `In Chapel Hill, what is the average salary ...`<br>1. In Chapel Hill, what is the average salary of residents with one house ?<br>2. In Chapel Hill, what is the average salary of residents with expenditure more than $60000 ? |

**(f) Auto-filler of questions based on user search and recommended questions by the system.**

**Figure 9: Exploratory analysis using the system**

eliminate such questions from the ranking process and subsequent recommendations, we have used GPT-2 [8] language model right after the generation process to verify the language correctness of questions. The questions generated by the system without any user feedback largely rely on the interestingness measure used and the ranking of these questions can be inaccurate but as the iterative exploration proceeds, the search space reduces leading to more relevant questions for the user.

We have performed an user evaluation study to understand the system's performance comprehensively. The questions generated were evaluated by three experts on three different criterias namely Fluency, Faithfulness, and Interestingness. We define them as follows.

| Fluency | Faithfulness | Interestingness |
|---|---|---|
| 4.8 | 4.6 | 4.1 |

**Table 3: Manual evaluation study**

**Fluency**: The question should be coherent without any grammatical errors.

**Faithfulness**: The question should be answerable using the information from the input table only. This ensures that the question is grounded on the input table.

**Interestingness**: The question should be about an interesting insight on the dataset. It captures how interesting is the generated question(inherently with answer).

Each question was scored from 1 to 5 with 1 being the worst rating and 5 being the best for each criteria, and the final score is an average across all the expert evaluators as shown in Table 3. Though this evaluation study is very limited, it shows the utility of the system to some extent. It motivates us to think about a promising direction of research on how to evaluate generated questions in a better way. Also We can enhance the system by considering more complex questions with a larger number of columns and operators. The choice of T5 language model is not a strict requirement as any other similar language model can be used too.

## 7 CONCLUSION

Conversational data exploration is gaining a lot of traction in the exploratory data analysis space of late. But there is limited work on automating the data exploration tasks. In this paper, we have taken a step towards building an end-to-end system for recommending interesting and relevant aggregate questions during conversational data exploration sessions. Firstly, we have created a custom dataset of interesting aggregate questions manually using a set of Kaggle datasets. We then used it for fine-tuning a pre-trained language model(T5) for question generation. We considered a select set of interestingness measures from the literature to identify interesting columns and data slices. This information is then used to slot-fill the questions generated by the fine-tuned language model. When the user starts exploration, the system recommendations are largely driven by the dataset but as the exploration proceeds, the user feedback helps to rank the recommendations better. We have experimented with this system on a variety of Kaggle datasets from different domains. The results show that the system generates interesting questions which will be very useful during exploratory data analysis. Thus our system uses fine-tuned language models along with interestingness measures and slot-filling to recommend interesting aggregate questions during exploratory analysis of structured datasets. In future, we plan to evaluate the system on more enterprise datasets. We also plan to experiment with other interestingness measures and slot-filling approaches to improve the system.

Automated Question Generation on Tabular Data for Conversational Data Exploration     Conference acronym 'XX, June 03–05, 2018, Woodstock, NY